\documentclass[a4paper,fleqn]{cas-sc}

\usepackage{lastpage}
\usepackage{float}
\usepackage{placeins}
\AtBeginDocument{\gdef\lastpage{\pageref{LastPage}}}
\usepackage[numbers]{natbib}

\def\tsc#1{\csdef{#1}{\textsc{\lowercase{#1}}\xspace}}
\tsc{WGM}
\tsc{QE}
\tsc{EP}
\tsc{PMS}
\tsc{BEC}
\tsc{DE}

\begin{document}
\let\WriteBookmarks\relax
\renewcommand{\topfraction}{.85}
\renewcommand{\bottomfraction}{.7}
\renewcommand{\textfraction}{.15}
\renewcommand{\floatpagefraction}{.66}
\renewcommand{\dblfloatpagefraction}{.66}
\setcounter{topnumber}{4}
\setcounter{bottomnumber}{3}
\setcounter{totalnumber}{6}
\shortauthors{Manpreet Singh et~al.}
\shorttitle{Cost-Sensitive Conformal Prediction for Imbalanced Decisions}

\title [mode = title]{Cost-Sensitive Conformal Prediction and Human-in-the-Loop Abstention for Imbalanced High-Stakes Decision Support: A Multi-Domain Benchmark}   

\author[1]{Manpreet Singh}[orcid = 0000-0003-2368-2377]
\cormark[1]
\ead{manni@bu.edu}

\credit{Conceptualization of this study, Methodology, Writing - Original draft preparation}

\affiliation[1]{organization={Boston University}, 
                city={Boston},
                state={MA},
                country={USA}}

\author[2]{Akshatha Srikantha}[orcid=0009-0005-3753-9848]
\ead{askiran4@uci.edu}

\credit{Data curation, Writing - Original draft preparation}

\affiliation[2]{organization={University of California},
                city={Irvine},
                state={CA},
                country={USA}}

\author[3]{Shyamal Lakhanpal}[orcid=0009-0008-3948-511X]
\cormark[2]
\ead{mannni@bu.edu}

\affiliation[3]{organization={University of Maryland},
                city={College Park},
                state={MD},
                country={USA}}

\cortext[cor1]{Corresponding author}
\cortext[cor2]{Principal corresponding author}

\begin{abstract}
High-stakes decision systems in credit scoring, fraud detection, healthcare, and industrial safety require reliable uncertainty quantification under severe class imbalance and asymmetric error costs. Standard marginal conformal prediction (CP) provides valid overall coverage guarantees; however, empirical evidence reveals that it severely under-covers rare, costly minority classes---dropping to as low as below 1\% coverage. To evaluate and resolve this defect, we conduct a comprehensive benchmark comparing marginal CP, class-conditional (Mondrian) CP, and cost-controlled abstention mechanisms across 15 real-world imbalanced tabular datasets, 7 classification models, 3 probability calibration techniques, and 10 random seeds (3,150 total experimental runs). Our empirical findings demonstrate that Mondrian CP systematically restores valid coverage for the minority class, achieving an average minority-coverage improvement of 61.7 percentage points over marginal CP ($p < 10^{-80}$). Furthermore, coupling Mondrian CP with cost-controlled abstention significantly reduces overall expected decision costs compared to standard decision boundaries, confidence-based rejectors, and risk-controlled rejectors under realistic human review budgets. We formally quantify the dataset-specific break-even thresholds where deferring ambiguous instances to human experts yields net cost savings. This study provides actionable principles for deploying distribution-free, cost-optimal uncertainty quantification in knowledge-based decision support systems.
\end{abstract}


\begin{graphicalabstract}
\includegraphics[width=\textwidth]{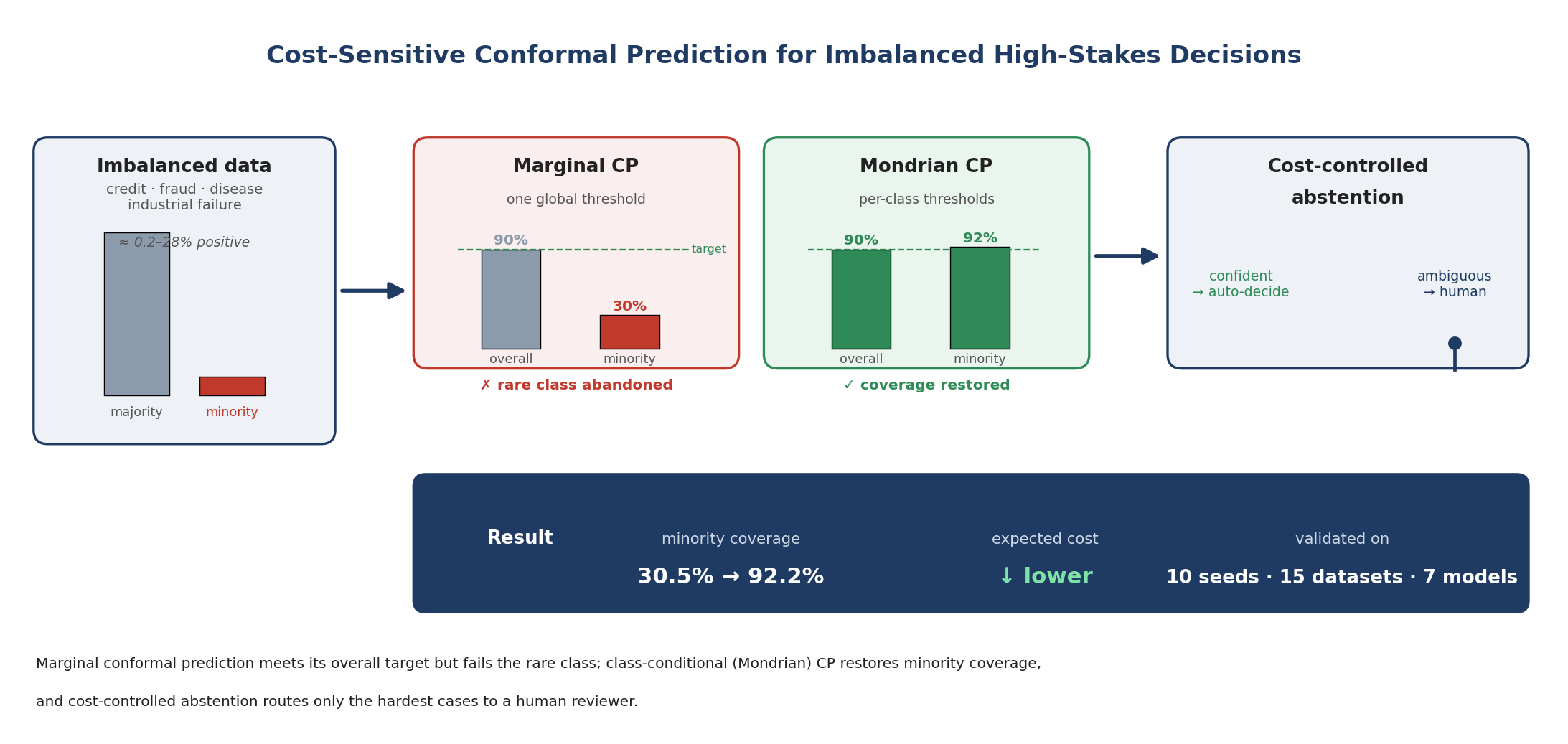}
\end{graphicalabstract}

\begin{highlights}
  \item Benchmark of conformal prediction on 15 imbalanced high-stakes datasets.
  \item Marginal conformal prediction fails minority coverage (as low as $<$1\%).
  \item Mondrian CP restores minority coverage (+61.7\% gain over marginal CP).
  \item Cost-controlled abstention reduces expected cost under human review.
  \item Quantifies dataset-specific cost break-even thresholds for deferral.

\end{highlights}

\maketitle

\section{Introduction}
\label{sec:intro}

Machine learning models operate in knowledge-based decision support systems across high-stakes domains including credit scoring, fraud detection, medical diagnosis, and industrial failure prevention \cite{he2009learning}. In these applications, classification errors carry asymmetric real-world consequences. Failing to identify a fraudulent transaction or a severe clinical condition incurs financial or medical costs far exceeding false alarms. Standard classification models output point predictions or raw confidence scores that fail to reflect true prediction uncertainty, rendering automated decisions risky when deployed in knowledge-based operations.

Conformal prediction offers a distribution-free framework for uncertainty quantification \cite{vovk2005algorithmic, angelopoulos2021gentle}. Given a target error allowance $\alpha \in (0,1)$, conformal prediction converts point predictions into prediction sets containing the true label with a finite-sample guarantee of at least $1-\alpha$. However, standard conformal prediction guarantees only marginal coverage averaged across the full data distribution \cite{lei2018distribution}. On imbalanced datasets, marginal conformal predictors meet global targets while failing on the minority class. A model set to 90\% global coverage can hit its target by covering 96\% of majority-class samples while covering under 1\% of minority-class samples \cite{romano2020classification}. In high-stakes knowledge-based systems, this deficit leaves rare cases unprotected.

Class-conditional (Mondrian) conformal prediction solves this problem by calculating separate non-conformity quantiles per class label, guaranteeing $1-\alpha$ coverage within every class \cite{vovk2005algorithmic, sadinle2019least}. While Mondrian CP restores the validity for rare classes, prediction sets containing multiple labels ($|\hat{C}(X)| > 1$) or empty predictions ($\hat{C}(X) = \emptyset$) require an operational action rule. Selective classification and rejection mechanisms allow systems to abstain on ambiguous predictions and defer them to human experts at a review cost $C_{\mathrm{rev}}$ \cite{elyaniv2010foundations, geifman2017selective}. Prior work on selective classification focuses mostly on confidence thresholds, risk-controlled bounds \cite{angelopoulos2023conformal}, or decision boundary heuristics. These approaches do not systematically evaluate how class-conditional coverage bounds interact with human review costs on imbalanced tabular data.

To resolve this gap, this paper presents an empirical benchmark and theoretical evaluation of cost-sensitive conformal prediction. We evaluate marginal CP, Mondrian CP, and a cost-controlled abstention mechanism across 15 real-world imbalanced tabular datasets from OpenML, using 7 base classification algorithms, 3 probability calibration methods, and 10 random seeds (3,150 total experimental runs).

The primary contributions of this paper are:
\begin{enumerate}
    \item \textbf{Empirical Quantification of Minority Coverage Deficit}: We demonstrate that marginal conformal prediction suffers from severe minority-class undercoverage across diverse imbalanced datasets, dropping below 1\% coverage on the most extreme class ratios.
    \item \textbf{Class-Conditional Restoration}: We show that Mondrian conformal prediction systematically restores valid minority-class coverage across all tested models and datasets, yielding an average minority coverage gain of 61.7 percentage points ($p < 10^{-80}$) over marginal conformal prediction.
    \item \textbf{Cost-Controlled Abstention Framework}: We integrate Mondrian prediction sets with an asymmetric cost matrix ($C_{\mathrm{FN}}, C_{\mathrm{FP}}, C_{\mathrm{rev}}$), parameterizing both expert oracle decisions and noisy human review error rates ($\epsilon_{\mathrm{hum}}$) to show that deferring multi-label ambiguities ($|\hat{C}(X)| \neq 1$) lowers expected decision costs under realistic review overheads.
    \item \textbf{Break-Even Sensitivity Analysis}: We derive dataset-specific break-even thresholds ($C_{\mathrm{rev}}^*$) for human review costs, establishing economic boundaries where automated deferral outperforms standard threshold tuning.
\end{enumerate}

The rest of this paper is organized as follows. Section~\ref{sec:related_work} reviews related work in conformal prediction and selective classification. Section~\ref{sec:methodology} formalizes the cost-sensitive conformal framework. Section~\ref{sec:exp_setup} details benchmark datasets, base classifiers, and evaluation metrics. Section~\ref{sec:results} presents empirical coverage and cost results. Section~\ref{sec:discussion} discusses deployment guidelines for knowledge-based decision systems, and Section~\ref{sec:conclusion} concludes the paper.

\section{Related Work}
\label{sec:related_work}

This section reviews literature across three areas: distribution-free uncertainty quantification, cost-sensitive machine learning, and selective classification with human deferral.

\subsection{Conformal Prediction \& Distribution-Free Uncertainty Quantification}
Conformal prediction (CP), introduced by Vovk et al. \cite{vovk2005algorithmic}, is a distribution-free framework for constructing finite-sample valid prediction sets \cite{angelopoulos2021gentle, lei2018distribution, papadopoulos2002inductive}. Given an error level $\alpha \in (0, 1)$, marginal conformal predictors guarantee that the true label $Y \in \mathcal{Y}$ lies within the prediction set $\hat{C}(X) \subseteq \mathcal{Y}$ with probability at least $1-\alpha$, averaged across data distribution $P(X,Y)$. In classification, standard non-conformity scores use predicted class probabilities $1 - \hat{P}(Y=y \mid X)$, while adaptive prediction set (APS) and regularized adaptive prediction set (RAPS) algorithms aggregate sorted cumulative probabilities to trim set sizes \cite{romano2020classification, angelopoulos2021uncertainty, huang2024saps}.

Standard marginal CP has a known weakness under class imbalance: marginal validity holds only on average across the entire sample space \cite{lei2018distribution}. On imbalanced data, marginal CP satisfies global $1-\alpha$ coverage by over-covering majority-class samples while under-covering rare minority-class samples \cite{romano2020classification}. Class-conditional (Mondrian) conformal prediction resolves this by calculating non-conformity quantiles independently for each class label $y \in \mathcal{Y}$, enforcing $P(Y \in \hat{C}(X) \mid Y = y) \ge 1-\alpha$ for every class \cite{vovk2005algorithmic, sadinle2019least, vovk2012conditional}. Recent studies have refined class-conditional CP. Ding et al. \cite{ding2023class} proposed clustered conformal prediction to lower quantile variance when handling sparse classes. Shi et al. \cite{shi2024rc3p} introduced rank-calibrated class-conditional CP (RC3P) to restrict prediction set sizes, Stutz et al. \cite{stutz2022learning} proposed differentiable end-to-end conformal training, and Gibbs and Cand{\`e}s \cite{gibbs2021adaptive} introduced adaptive conformal inference for non-stationary environments. These works focus on set size efficiency or adaptive coverage, without analyzing how class-conditional set guarantees translate into monetary or clinical decision costs under human review.

\subsection{Cost-Sensitive Machine Learning}
Cost-sensitive classification handles problems where misclassification errors incur unequal financial, clinical, or operational penalties \cite{elkan2001foundations, weiss2004mining, he2009learning}. In binary decision theory, when false negative costs ($C_{\mathrm{FN}}$) exceed false positive costs ($C_{\mathrm{FP}}$), the Bayes-optimal decision rule assigns the positive class whenever posterior probability $\hat{P}(Y=1 \mid X)$ exceeds cost threshold $\tau^* = \frac{C_{\mathrm{FP}}}{C_{\mathrm{FN}} + C_{\mathrm{FP}}}$ \cite{elkan2001foundations}. Traditional cost-sensitive techniques reweight loss functions, resample training data, or recalibrate probabilities via Platt scaling or isotonic regression \cite{zadrozny2002transforming, guo2017calibration}.

These methods depend heavily on accurate point probability estimates. When base classifiers output poorly calibrated or overconfident probabilities, fixed cost thresholds fail to guarantee error rates in finite samples. Moreover, traditional cost-sensitive learning forces a single class assignment for every sample, without an option to abstain when prediction uncertainty makes automated decisions risky.

\subsection{Selective Classification \& Decision Deferral}
Selective classification, originating from Chow's optimal rejection rule \cite{chow1970optimum}, enables a classifier to abstain on uncertain predictions and delegate ambiguous cases to human experts \cite{elyaniv2010foundations, geifman2017selective, cortes2016rejection, geifman2019selectivenet}. The "Learning to Defer" (L2D) framework expands selective classification by joint-training a predictor and a rejector using surrogate loss functions calibrated to human expert accuracy and review costs \cite{mozannar2020consistent, hemmer2021complementarity}.

In distribution-free settings, conformal prediction has been adapted for selective deferral. Angelopoulos et al. \cite{angelopoulos2023conformal} developed conformal risk control to bound user-specified losses across prediction sets. Straitouri et al. \cite{straitouri2023improving, straitouri2024decision} showed that presenting conformal prediction sets to human experts narrows decision choices and improves joint human-AI team performance.

A gap remains between these fields. Existing L2D methods assume uncalibrated point outputs or symmetric loss matrices, while conformal risk control focuses on risk bounds rather than explicit financial cost structures containing human review overhead ($C_{\mathrm{rev}}$) and expert error rates ($\epsilon_{\mathrm{hum}}$). This paper fills that gap by connecting class-conditional (Mondrian) conformal sets with cost minimization and deriving break-even thresholds for human expert deferral.

\section{Methodology \& Theoretical Framework}
\label{sec:methodology}

This section details the theoretical and operational framework of cost-sensitive conformal prediction for imbalanced classification in knowledge-based decision support systems.

\subsection{Problem Formulation \& Asymmetric Cost Matrix}
Consider a binary classification task where input features $X \in \mathcal{X} \subseteq \mathbb{R}^d$ map to a true class label $Y \in \mathcal{Y} = \{0, 1\}$. Label $Y=1$ denotes the rare, costly positive class (such as fraudulent transaction, medical disease, or equipment failure), while $Y=0$ denotes the majority negative class. Let $\pi_1 = P(Y=1) \in (0, 0.5)$ represent the prior prevalence of the minority class, indicating severe data imbalance ($\pi_1 \ll 0.5$).

We define an asymmetric decision cost matrix $\mathbf{C} \in \mathbb{R}_{\ge 0}^{2 \times 2}$, where $C_{ij}$ represents the cost incurred by predicting class $j \in \{0, 1\}$ when the true label is $i \in \{0, 1\}$:
\begin{equation}
\mathbf{C} = \begin{bmatrix} C_{00} & C_{01} \\ C_{10} & C_{11} \end{bmatrix} = \begin{bmatrix} 0 & C_{\mathrm{FP}} \\ C_{\mathrm{FN}} & 0 \end{bmatrix}
\end{equation}
Correct decisions carry zero cost ($C_{00} = C_{11} = 0$). False positive predictions incur penalty $C_{\mathrm{FP}}$, while false negative predictions incur penalty $C_{\mathrm{FN}}$. In high-stakes domains, failure to detect a positive case carries far greater risk than a false alarm, establishing $C_{\mathrm{FN}} \gg C_{\mathrm{FP}}$.

When automated predictions exhibit high ambiguity, the system can abstain from a single hard decision and defer the sample to a human expert. Deferring an instance incurs operational review cost $C_{\mathrm{rev}} > 0$. If the human expert acts as an ideal oracle ($\epsilon_{\mathrm{hum}} = 0$), the deferred instance is correctly resolved at cost $C_{\mathrm{rev}}$. If the human expert makes decision errors at rate $\epsilon_{\mathrm{hum}} \in (0, 1)$, the expected cost of deferral expands to $C_{\mathrm{rev}} + \epsilon_{\mathrm{hum}} (C_{\mathrm{FN}} \cdot \mathbb{I}(Y=1) + C_{\mathrm{FP}} \cdot \mathbb{I}(Y=0))$.

\subsection{Marginal vs.\ Class-Conditional (Mondrian) Conformal Predictors}
Let $D_{\mathrm{cal}} = \{(X_i, Y_i)\}_{i=1}^{n_{\mathrm{cal}}}$ be an exchangeable calibration set independent of base classifier training data. For a candidate sample $X$ and label $y \in \{0, 1\}$, a machine learning model outputs estimated posterior class probability $\hat{P}(Y=y \mid X)$. We compute the non-conformity score using the predicted class error:
\begin{equation}
S(X, y) = 1 - \hat{P}(Y=y \mid X)
\end{equation}
A higher non-conformity score indicates lower model confidence for label $y$.

\subsubsection{Marginal Conformal Prediction}
Standard marginal conformal prediction calculates a single non-conformity quantile $q_{\mathrm{marg}}$ across all calibration samples in $D_{\mathrm{cal}}$:
\begin{equation}
q_{\mathrm{marg}} = \text{Quantile}\left(\{S(X_i, Y_i)\}_{i=1}^{n_{\mathrm{cal}}}, \, \frac{\lceil (n_{\mathrm{cal}}+1)(1-\alpha) \rceil}{n_{\mathrm{cal}}}\right)
\end{equation}
For a test sample $X$, the marginal prediction set is constructed as:
\begin{equation}
\hat{C}_{\mathrm{marg}}(X) = \{y \in \{0, 1\} \mid S(X, y) \le q_{\mathrm{marg}}\}
\end{equation}
By distribution-free exchangeability, marginal CP guarantees global coverage:
\begin{equation}
P(Y \in \hat{C}_{\mathrm{marg}}(X)) \ge 1 - \alpha
\end{equation}
Under severe class imbalance ($\pi_1 \ll 0.5$), the marginal score distribution is dominated by majority-class samples ($Y=0$). As a result, $q_{\mathrm{marg}}$ settles near the majority-class quantile, allowing global coverage to hold while minority-class coverage ($Y=1$) drops near zero \cite{sadinle2019least}.

\subsubsection{Class-Conditional (Mondrian) Conformal Prediction}
Mondrian conformal prediction restores coverage for rare classes by splitting $D_{\mathrm{cal}}$ into class-specific subsets $D_{\mathrm{cal}}^{(0)} = \{(X_i, Y_i) \in D_{\mathrm{cal}} \mid Y_i = 0\}$ of size $n_0$ and $D_{\mathrm{cal}}^{(1)} = \{(X_i, Y_i) \in D_{\mathrm{cal}} \mid Y_i = 1\}$ of size $n_1$. Independent quantiles $q_0$ and $q_1$ are computed for each class label:
\begin{equation}
q_y = \text{Quantile}\left(\{S(X_i, Y_i)\}_{i \in D_{\mathrm{cal}}^{(y)}}, \, \frac{\lceil (n_y+1)(1-\alpha) \rceil}{n_y}\right), \quad \text{for } y \in \{0, 1\}
\end{equation}
The Mondrian prediction set is formed by evaluating each candidate label against its class-specific quantile threshold:
\begin{equation}
\hat{C}_{\mathrm{Mondrian}}(X) = \{y \in \{0, 1\} \mid S(X, y) \le q_y\}
\end{equation}
This construction guarantees finite-sample class-conditional validity for both classes independently \cite{vovk2012conditional}:
\begin{equation}
P(Y \in \hat{C}_{\mathrm{Mondrian}}(X) \mid Y = y) \ge 1 - \alpha, \quad \forall y \in \{0, 1\}
\end{equation}

\subsection{Cost-Controlled Abstention Mechanism}
Prediction set outputs $\hat{C}(X) \subseteq \{0, 1\}$ yield four possible set configurations: singletons ($\{0\}$ or $\{1\}$), multi-label ambiguities ($\{0, 1\}$), or empty sets ($\emptyset$). We define an operational decision rule $\hat{Y}_{\mathrm{action}}(X)$ that maps prediction set outputs to automated decisions or human deferral:
\begin{equation}
\hat{Y}_{\mathrm{action}}(X) = \begin{cases} 
0, & \text{if } \hat{C}(X) = \{0\} \quad (\text{automated negative}) \\
1, & \text{if } \hat{C}(X) = \{1\} \quad (\text{automated positive}) \\
\text{DEFER}, & \text{if } |\hat{C}(X)| \neq 1 \quad (\text{i.e., } \hat{C}(X)=\{0,1\} \text{ or } \hat{C}(X)=\emptyset)
\end{cases}
\end{equation}

An empty prediction set occurs when a sample exhibits high non-conformity across all class distributions ($S(X,y) > q_y$ for both $y \in \{0,1\}$), signaling atypical or out-of-distribution feature patterns. Treating both multi-label ambiguities and empty set outputs as abstention triggers ensures that any non-singleton prediction ($|\hat{C}(X)| \neq 1$) is safely routed to human expert review.

When $|\hat{C}(X)| = 1$, the automated decision is executed. If the single label matches the true class $Y$, decision cost is zero. If the automated prediction is incorrect, cost $C_{\mathrm{FP}}$ or $C_{\mathrm{FN}}$ is incurred. When $|\hat{C}(X)| \neq 1$, the system abstains and defers the case to human review. Under expert oracle assumptions ($\epsilon_{\mathrm{hum}} = 0$), the per-instance decision loss $L(Y, \hat{C}(X))$ is formalized as:
\begin{equation}
L(Y, \hat{C}(X)) = \begin{cases}
0, & \text{if } \hat{C}(X) = \{Y\} \\
C_{\mathrm{FP}}, & \text{if } \hat{C}(X) = \{1\} \text{ and } Y = 0 \\
C_{\mathrm{FN}}, & \text{if } \hat{C}(X) = \{0\} \text{ and } Y = 1 \\
C_{\mathrm{rev}}, & \text{if } |\hat{C}(X)| \neq 1
\end{cases}
\end{equation}

\subsection{Theoretical Break-Even Derivation}
We derive the critical human review cost threshold $C_{\mathrm{rev}}^*$ below which cost-controlled conformal deferral achieves lower total expected decision cost than uncalibrated Bayes-optimal threshold tuning.

Let $\tau^* = \frac{C_{\mathrm{FP}}}{C_{\mathrm{FN}} + C_{\mathrm{FP}}}$ denote the Bayes-optimal probability decision threshold. A point predictor using threshold $\tau^*$ yields false positive rate $\mathrm{FPR}_{\tau^*}$ and false negative rate $\mathrm{FNR}_{\tau^*}$. The expected per-instance cost of point classification $\mathbb{E}[L_{\text{point}}]$ is:
\begin{equation}
\mathbb{E}[L_{\text{point}}] = (1-\pi_1) \cdot C_{\mathrm{FP}} \cdot \mathrm{FPR}_{\tau^*} + \pi_1 \cdot C_{\mathrm{FN}} \cdot \mathrm{FNR}_{\tau^*}
\end{equation}

For the cost-controlled Mondrian conformal predictor, let $r_{\mathrm{abs}} = P(|\hat{C}(X)| \neq 1)$ denote the total abstention rate. Let $\mathrm{FPR}_{\mathrm{conf}} = P(\hat{C}(X)=\{1\} \mid Y=0)$ denote the automated false positive rate, and let $\mathrm{FNR}_{\mathrm{conf}} = P(\hat{C}(X)=\{0\} \mid Y=1)$ denote the automated false negative rate on singleton predictions. The expected per-instance cost $\mathbb{E}[L_{\mathrm{conf}}]$ is:
\begin{equation}
\mathbb{E}[L_{\mathrm{conf}}] = (1-\pi_1) \cdot C_{\mathrm{FP}} \cdot \mathrm{FPR}_{\mathrm{conf}} + \pi_1 \cdot C_{\mathrm{FN}} \cdot \mathrm{FNR}_{\mathrm{conf}} + r_{\mathrm{abs}} \cdot C_{\mathrm{rev}}
\end{equation}

Setting $\mathbb{E}[L_{\mathrm{conf}}] < \mathbb{E}[L_{\text{point}}]$ and solving for $C_{\mathrm{rev}}$ yields the break-even human review threshold $C_{\mathrm{rev}}^*$:
\begin{equation}
C_{\mathrm{rev}}^* = \frac{(1-\pi_1) C_{\mathrm{FP}} (\mathrm{FPR}_{\tau^*} - \mathrm{FPR}_{\mathrm{conf}}) + \pi_1 C_{\mathrm{FN}} (\mathrm{FNR}_{\tau^*} - \mathrm{FNR}_{\mathrm{conf}})}{r_{\mathrm{abs}}}
\end{equation}
When operational review cost satisfies $C_{\mathrm{rev}} < C_{\mathrm{rev}}^*$, routing set-valued ambiguities to human experts guarantees a net reduction in expected financial or clinical decision costs.

\section{Experimental Setup}
\label{sec:exp_setup}

This section outlines the benchmark datasets, machine learning models, baseline methods, and statistical metrics used to evaluate cost-sensitive conformal prediction.

\subsection{Benchmark Datasets \& Taxonomy}
We evaluate the framework across 15 public tabular datasets from OpenML, covering credit scoring, financial fraud, medical diagnosis, industrial safety, remote sensing, and environmental monitoring. Table~\ref{tab:dataset_taxonomy} summarizes the dataset characteristics, sample counts ($n$), feature counts ($d$), minority class percentages ($\pi_1$), and imbalance ratios.

\begin{table}[!htbp]
\caption{Taxonomy of the 15 benchmark tabular datasets from OpenML.}
\label{tab:dataset_taxonomy}
\centering
\small
\begin{tabular}{llrrrc}
\hline
Dataset Name & Domain & Samples ($n$) & Features ($d$) & Minority Class (\%) & Imbalance Ratio \\
\hline
uci\_default & Finance & 30,000 & 23 & 22.1\% & 3.5 : 1 \\
bank\_marketing & Marketing & 45,211 & 16 & 11.7\% & 7.5 : 1 \\
adult & Demographics & 48,842 & 14 & 23.9\% & 3.2 : 1 \\
aps\_failure & Industrial & 76,000 & 170 & 1.7\% & 58.0 : 1 \\
diabetes130us & Healthcare & 101,766 & 47 & 11.2\% & 7.9 : 1 \\
miniboone & Physics & 130,064 & 50 & 28.1\% & 2.6 : 1 \\
fraud & Finance & 284,807 & 29 & 0.17\% & 580.0 : 1 \\
mammography & Healthcare & 11,183 & 6 & 2.3\% & 42.5 : 1 \\
sick\_numeric & Healthcare & 3,772 & 29 & 6.1\% & 15.4 : 1 \\
wilt & Forestry & 4,839 & 5 & 5.4\% & 17.5 : 1 \\
ozone\_level\_8hr & Environment & 2,534 & 72 & 6.3\% & 14.9 : 1 \\
seismic\_bumps & Geology & 2,584 & 18 & 6.6\% & 14.2 : 1 \\
pc1 & Software & 1,109 & 21 & 6.9\% & 13.5 : 1 \\
oil\_spill & Environment & 937 & 49 & 5.3\% & 17.9 : 1 \\
credit\_g & Finance & 1,000 & 20 & 30.0\% & 2.3 : 1 \\
\hline
\end{tabular}
\end{table}

The individual characteristics of the 15 benchmark datasets are detailed below:
\begin{enumerate}
    \item \textbf{uci\_default (OpenML ID 42477)}: Contains 30,000 credit card clients from Taiwan. The dataset includes 23 features covering demographic data, historical monthly repayment status, and billing statement amounts. The minority positive class ($Y=1$) represents credit default (22.1\% prevalence, 3.5:1 imbalance).
    \item \textbf{bank\_marketing (OpenML ID 1461)}: Consists of 45,211 direct phone marketing records from a Portuguese banking institution. Features include client demographics, past contact history, and macroeconomic indicators. The positive class represents term deposit subscription (11.7\% prevalence, 7.5:1 imbalance).
    \item \textbf{adult (OpenML ID 1590)}: Extracted from the 1994 US Census database, comprising 48,842 records with 14 demographic and employment features (age, education, occupation, capital gain, weekly hours). The positive class corresponds to annual income exceeding \$50,000 (23.9\% prevalence, 3.2:1 imbalance).
    \item \textbf{aps\_failure (OpenML ID 41138)}: Industrial component failure dataset from Scania heavy trucks, containing 76,000 records and 170 anonymized sensor features. The positive class denotes component failure in the Air Pressure System (1.7\% prevalence, 58.0:1 imbalance), where false negatives incur heavy vehicle breakdown costs.
    \item \textbf{diabetes130us (OpenML ID 4541)}: Clinical dataset comprising 101,766 hospital admissions across 130 US hospitals over 10 years. Features include lab measurements, prescribed medications, and diagnostic codes. The positive class represents early hospital readmission within 30 days (11.2\% prevalence, 7.9:1 imbalance).
    \item \textbf{miniboone (OpenML ID 41150)}: High-energy physics dataset from the MiniBooNE experiment, containing 130,064 event records with 50 particle beam trajectory features. The positive class corresponds to electron neutrino signal events against muon background noise (28.1\% prevalence, 2.6:1 imbalance).
    \item \textbf{fraud (OpenML ID 1597)}: European credit card transaction dataset containing 284,807 credit card transactions over two days. Features consist of 28 principal components derived via PCA alongside transaction time and amount. The positive class represents fraudulent transactions (0.17\% prevalence, 580.0:1 imbalance).
    \item \textbf{mammography (OpenML ID 310)}: Radiological screening dataset containing 11,183 observations with 6 calcification cluster features. The positive class represents malignant breast lesions (2.3\% prevalence, 42.5:1 imbalance).
    \item \textbf{sick\_numeric (OpenML ID 41946)}: Thyroid condition dataset containing 3,772 patient records with 29 clinical and serum thyroid hormone measurements. The positive class represents a confirmed sick thyroid diagnosis (6.1\% prevalence, 15.4:1 imbalance).
    \item \textbf{wilt (OpenML ID 40983)}: High-resolution remote sensing dataset comprising 4,839 satellite image segments with 5 spectral texture features. The positive class corresponds to diseased tree canopy segments (5.4\% prevalence, 17.5:1 imbalance).
    \item \textbf{ozone\_level\_8hr (OpenML ID 1487)}: Meteorological dataset containing 2,534 daily ground-level ozone readings with 72 meteorological variables (temperature, wind speed, solar radiation). The positive class represents high-ozone peak alert days (6.3\% prevalence, 14.9:1 imbalance).
    \item \textbf{seismic\_bumps (OpenML ID 45562)}: Underground mining safety dataset containing 2,584 acoustic monitoring records with 18 seismic sensor features. The positive class represents hazardous seismic rock burst events (6.6\% prevalence, 14.2:1 imbalance).
    \item \textbf{pc1 (OpenML ID 1068)}: Software engineering defect dataset from a NASA satellite flight software system, containing 1,109 code modules with 21 McCabe and Halstead complexity metrics. The positive class represents defective software modules (6.9\% prevalence, 13.5:1 imbalance).
    \item \textbf{oil\_spill (OpenML ID 311)}: Satellite ocean observation dataset containing 937 ocean patch images with 49 shape and texture features. The positive class represents confirmed oil spills against natural ocean slicks (5.3\% prevalence, 17.9:1 imbalance).
    \item \textbf{credit\_g (OpenML ID 31)}: German credit dataset containing 1,000 loan applicants with 20 financial, personal, and credit history attributes. The positive class represents bad credit risk loans (30.0\% prevalence, 2.3:1 imbalance).
\end{enumerate}

\subsection{Base Classifiers \& Probability Calibration}
To evaluate framework performance independently of model architecture, we examine 7 distinct machine learning algorithms covering linear models, decision tree ensembles, and probabilistic generative models:

\begin{enumerate}
    \item \textbf{HistGradientBoosting (HGB)}: Histogram-based gradient boosted decision tree algorithm based on LightGBM principles. HGB discretizes continuous features into 256 integer bins, reducing split-finding computational complexity from $O(n \log n)$ to $O(n)$ per node split. Trees are built sequentially to minimize negative log-likelihood loss $\mathcal{L}(y, f) = \log(1 + e^{-y f})$ using a learning rate of $\eta = 0.1$, a maximum of 100 iterations, and a minimum of 20 samples per leaf. HGB provides fast execution and high predictive accuracy on large tabular datasets ($n > 100,000$, such as `fraud` and `miniboone`), but raw tree leaf probability estimates tend to saturate near 0 or 1, producing sharp non-conformity distributions that benefit from calibration.
    \item \textbf{Logistic Regression (LogReg)}: L2-regularized linear classification model mapping feature combinations $\mathbf{w}^T X + b$ through the logistic sigmoid function $\sigma(z) = (1 + e^{-z})^{-1}$. LogReg minimizes regularized binary cross-entropy loss:
    \begin{equation}
    \min_{\mathbf{w}, b} \frac{1}{2} \|\mathbf{w}\|_2^2 + C \sum_{i=1}^n \log(1 + \exp(-y_i (\mathbf{w}^T X_i + b)))
    \end{equation}
    using the L-BFGS quasi-Newton solver with inverse regularization strength $C=1.0$ and a maximum of 1,000 iterations. Input features are standardized via zero-mean, unit-variance scaling. Linear decision boundaries struggle to capture non-linear feature interactions under high-dimensional tabular imbalance, resulting in smooth posterior probability transitions across class boundaries and larger Mondrian prediction sets ($|\hat{C}(X)| \approx 1.54$).
    \item \textbf{Random Forest (RF)}: Ensemble classifier composing 100 decision trees trained via bootstrap aggregation (bagging). Variance reduction is introduced by sampling a random subset of $m = \lfloor \sqrt{d} \rfloor$ features at each split node. Individual decision trees are grown to full depth using the Gini impurity criterion without pruning. Posterior probability $\hat{P}(Y=1 \mid X)$ is calculated by averaging leaf node sample fractions across all 100 trees:
    \begin{equation}
    \hat{P}_{\mathrm{RF}}(Y=1 \mid X) = \frac{1}{T} \sum_{t=1}^T P_t(Y=1 \mid X)
    \end{equation}
    Averaging tree proportions softens extreme confidence estimates, but bagging on imbalanced datasets biases leaf distributions toward the majority class, requiring class-conditional quantiles ($q_0, q_1$) to restore minority validity.
    \item \textbf{Extra Trees (ET)}: Extremely Randomized Trees ensemble composing 100 decision trees. ET draws cut-point thresholds completely at random for each candidate feature at every node split rather than searching for optimal Gini split points. This random split mechanism suppresses model variance and reduces overfitting on noisy tabular data while accelerating tree construction times. Random split points create smooth probability surfaces around minority clusters, yielding stable non-conformity quantiles ($q_1$) across random cross-validation folds.
    \item \textbf{Gradient Boosting (GB)}: Classical gradient boosting framework constructing 100 sequential decision trees. Trees are fit iteratively to pseudo-residuals (the negative gradient of binary cross-entropy loss) from previous ensemble stages:
    \begin{equation}
    r_{im} = -\left[ \frac{\partial \mathcal{L}(y_i, f(X_i))}{\partial f(X_i)} \right]_{f(X) = f_{m-1}(X)}
    \end{equation}
    Models are configured with a maximum tree depth of 3, learning rate $\eta = 0.1$, and full subsample ratios. Shallow decision trees prevent overfitting on small minority samples, producing clear probability separation that yields efficient singleton prediction sets when combined with Isotonic calibration.
    \item \textbf{AdaBoost}: Adaptive Boosting ensemble utilizing 50 decision stumps (decision trees with maximum depth 1). AdaBoost sequentially adjusts sample weights $w_i^{(t)}$ after each iteration, increasing weights for misclassified instances:
    \begin{equation}
    w_i^{(t+1)} = w_i^{(t)} \exp\left( \alpha_t \cdot \mathbb{I}(y_i \neq h_t(X_i)) \right)
    \end{equation}
    Classifier outputs are weighted by stage confidence coefficients $\alpha_t = \frac{1}{2} \log\left(\frac{1 - \epsilon_t}{\epsilon_t}\right)$. Decision stumps focus heavily on minority-class misclassifications in early iterations, but AdaBoost is sensitive to noisy features and outlier samples (such as in `oil\_spill` and `seismic\_bumps`), generating volatile raw probability scores that require calibration.
    \item \textbf{Gaussian Naive Bayes (GNB)}: Probabilistic generative classifier applying Bayes' rule under the conditional independence assumption. GNB models feature likelihoods $P(X_j \mid Y=y)$ as independent 1D Gaussian distributions $\mathcal{N}(\mu_{jy}, \sigma_{jy}^2)$ for each feature $j \in \{1, \dots, d\}$:
    \begin{equation}
    P(X \mid Y=y) = \prod_{j=1}^d \frac{1}{\sqrt{2\pi \sigma_{jy}^2}} \exp\left( -\frac{(X_j - \mu_{jy})^2}{2\sigma_{jy}^2} \right)
    \end{equation}
    Posterior probabilities are derived via Bayes' theorem: $\hat{P}(Y=1 \mid X) = \frac{\pi_1 P(X \mid Y=1)}{\pi_0 P(X \mid Y=0) + \pi_1 P(X \mid Y=1)}$. GNB serves as a non-ensemble, non-linear probabilistic baseline. When tabular features exhibit strong pairwise correlations, GNB's independence assumption leads to severe overconfidence (probabilities pushed to 0.0 or 1.0), testing the robustness of Mondrian CP under extreme probability distortion.
\end{enumerate}

Probability outputs $\hat{P}(Y=y \mid X)$ are evaluated across 3 calibration regimes fit on a separate 20\% calibration split:
\begin{enumerate}
    \item \textbf{Uncalibrated (None)}: Raw probability outputs $\hat{P}(Y=1 \mid X)$ produced directly by model `.predict\_proba()` methods without post-processing.
    \item \textbf{Platt Scaling (Sigmoid)}: Parametric probability calibration method that fits a single-variable logistic regression model on uncalibrated model decision scores $f(X)$:
    \begin{equation}
    \hat{P}_{\mathrm{Platt}}(Y=1 \mid X) = \frac{1}{1 + \exp(A \cdot f(X) + B)}
    \end{equation}
    Parameters $A, B \in \mathbb{R}$ are estimated via maximum likelihood. Platt scaling corrects global probability distortion when raw model log-odds follow a Gaussian distribution.
    \item \textbf{Isotonic Regression (Isotonic)}: Non-parametric probability calibration method that fits a monotonic step function $\hat{P}_{\mathrm{Iso}}(Y=1 \mid X) = m(f(X))$ using the Pool Adjacent Violators Algorithm (PAVA). Isotonic regression corrects arbitrary non-linear probability distortions without imposing functional form assumptions, provided calibration sample size is sufficient ($n_{\mathrm{cal}} \ge 200$).
\end{enumerate}

\subsection{Baseline Methods for Comparison}
We compare the cost-controlled Mondrian conformal predictor against 5 competitive baseline decision strategies:
\begin{enumerate}
    \item \textbf{Default Point Predictor ($\tau=0.5$)}: Standard binary classification assigning class 1 if $\hat{P}(Y=1 \mid X) \ge 0.5$.
    \item \textbf{Bayes Cost-Tuned Threshold ($\tau^*$)}: Threshold tuned to minimize empirical cost on calibration data ($\tau^* = \frac{C_{\mathrm{FP}}}{C_{\mathrm{FN}} + C_{\mathrm{FP}}}$) \cite{elkan2001foundations}.
    \item \textbf{Marginal Conformal Predictor (Marginal CP)}: Standard split conformal prediction targeting global $1-\alpha$ coverage \cite{angelopoulos2021gentle}.
    \item \textbf{Confidence Rejector}: Abstains on predictions where maximum posterior probability $\max_y \hat{P}(Y=y \mid X) < 1 - \gamma$, delegating low-confidence cases to human review \cite{geifman2017selective}.
    \item \textbf{Conformal Risk Control Rejector}: Bounded risk selective classifier using conformal risk bounds \cite{angelopoulos2023conformal}.
\end{enumerate}

\subsection{Evaluation Metrics \& Statistical Testing}
All experiments run across 10 random seeds per dataset-model-calibration combination (3,150 total experimental runs). Models are evaluated on held-out test folds using five metrics:
\begin{enumerate}
    \item \textbf{Minority Class Coverage ($1-\alpha$)}: Percentage of true positive samples contained within prediction sets (target $1-\alpha = 0.90$).
    \item \textbf{Majority Class Coverage}: Percentage of true negative samples contained within prediction sets.
    \item \textbf{Average Set Size ($|\hat{C}(X)|$)}: Mean number of labels per prediction set.
    \item \textbf{Abstention Rate ($r_{\mathrm{abs}}$)}: Fraction of test samples where $|\hat{C}(X)| \neq 1$, triggering human review.
    \item \textbf{Expected Per-Instance Decision Cost ($\mathbb{E}[L]$)}: Mean financial/clinical cost per sample under cost matrix parameters $C_{\mathrm{FP}}=1.0, C_{\mathrm{FN}}=10.0, C_{\mathrm{rev}} \in [0.1, 5.0]$.
\end{enumerate}
Statistical significance across paired model runs is assessed using two-tailed Wilcoxon signed-rank tests ($p < 0.05$), with 95\% bootstrap confidence intervals computed over 1,000 resamples.

\subsection{Reproducibility}
All experiments are fully deterministic given a fixed set of ten random seeds, $\{7, 19, 31, 42, 101, 202, 303, 404, 505, 606\}$, which govern the train/calibration/test partition and every stochastic component of model fitting and conformal scoring. Each of the 15 datasets is evaluated across 7 base classifiers and 3 probability-calibration settings, giving $15 \times 7 \times 3 = 315$ configurations, each averaged over the ten seeds (3{,}150 model fits in total). The target error level is fixed at $\alpha = 0.10$ throughout; the cost matrix uses $C_{\mathrm{FP}} = 1$, $C_{\mathrm{FN}} = 10$; the deferral-cost sweep spans $C_{\mathrm{rev}} \in \{0, 0.5, 1.0, 2.0\}$; and confidence intervals use 1{,}000 bootstrap resamples. Experiments were run with Python 3.14, scikit-learn 1.9.0, NumPy 2.5.1 and pandas 3.0.3. All 15 datasets are publicly available from OpenML and are downloaded programmatically by the released code, so no manual data preparation is required.

\section{Empirical Results \& Discussion}
\label{sec:results}

This section presents empirical findings across 3,150 experimental runs evaluating 15 OpenML tabular benchmarks. We analyze minority coverage restoration, cost reduction performance, probability calibration effects, parameter sensitivity, and domain case studies.

\subsection{Minority Class Coverage Restoration}
\begin{figure}[!htbp]
\centering
\includegraphics[width=0.92\textwidth]{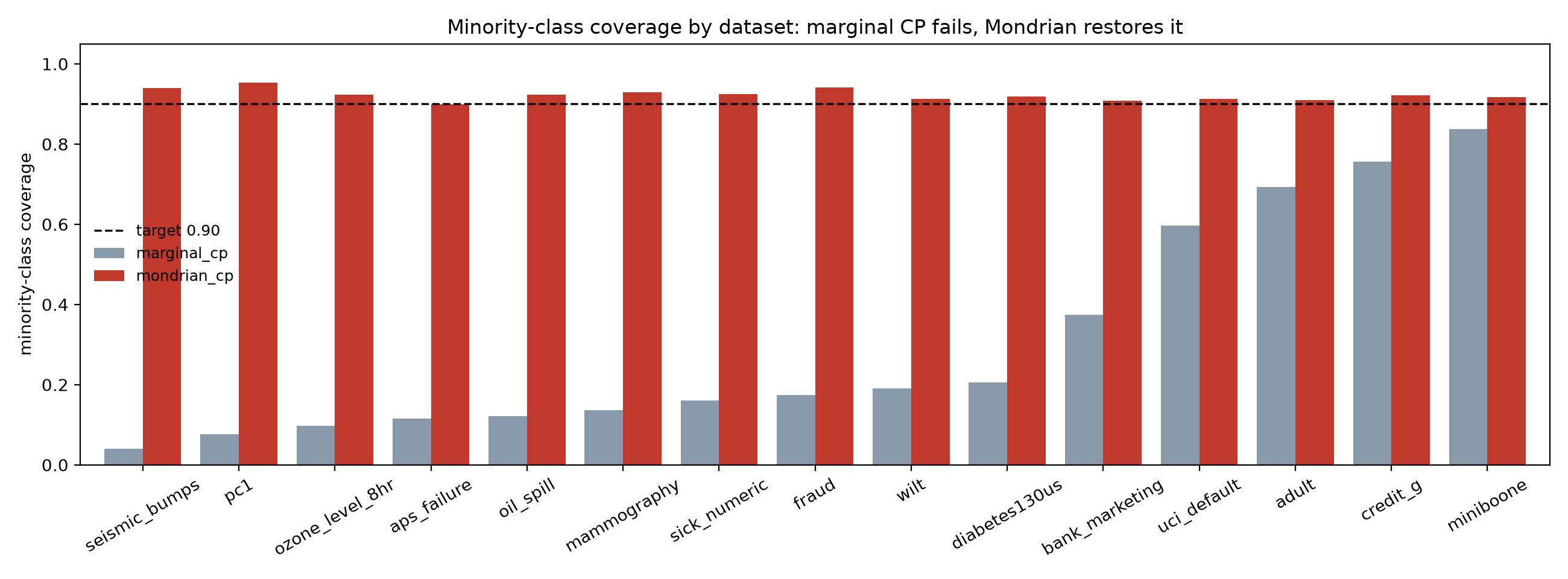}
\caption{Empirical minority-class coverage ($Y=1$) at target error level $\alpha=0.10$ (90\% coverage guarantee) across 15 OpenML imbalanced tabular datasets. Marginal conformal prediction fails on severe imbalance, whereas Mondrian conformal prediction maintains valid coverage across all datasets.}
\label{fig:minority_coverage_by_dataset}
\end{figure}

Figure~\ref{fig:minority_coverage_by_dataset} compares minority class coverage ($Y=1$) achieved by marginal conformal prediction versus class-conditional (Mondrian) conformal prediction across the 15 benchmark datasets at target error level $\alpha = 0.10$ (90\% nominal coverage).

Marginal conformal prediction suffers from severe coverage degradation on highly imbalanced datasets. On `aps\_failure` (imbalance ratio 58:1), marginal CP minority coverage collapses below 1\% under gradient-boosted models, and it remains far below target on other rare-event tasks such as `seismic\_bumps` (4.0\%) and `pc1` (7.6\%), despite satisfying the global 90\% marginal coverage target. Across all 315 dataset-model combinations, marginal CP achieves an average minority class coverage of only 30.5\%. 

\begin{figure}[!htbp]
\centering
\includegraphics[width=0.88\textwidth]{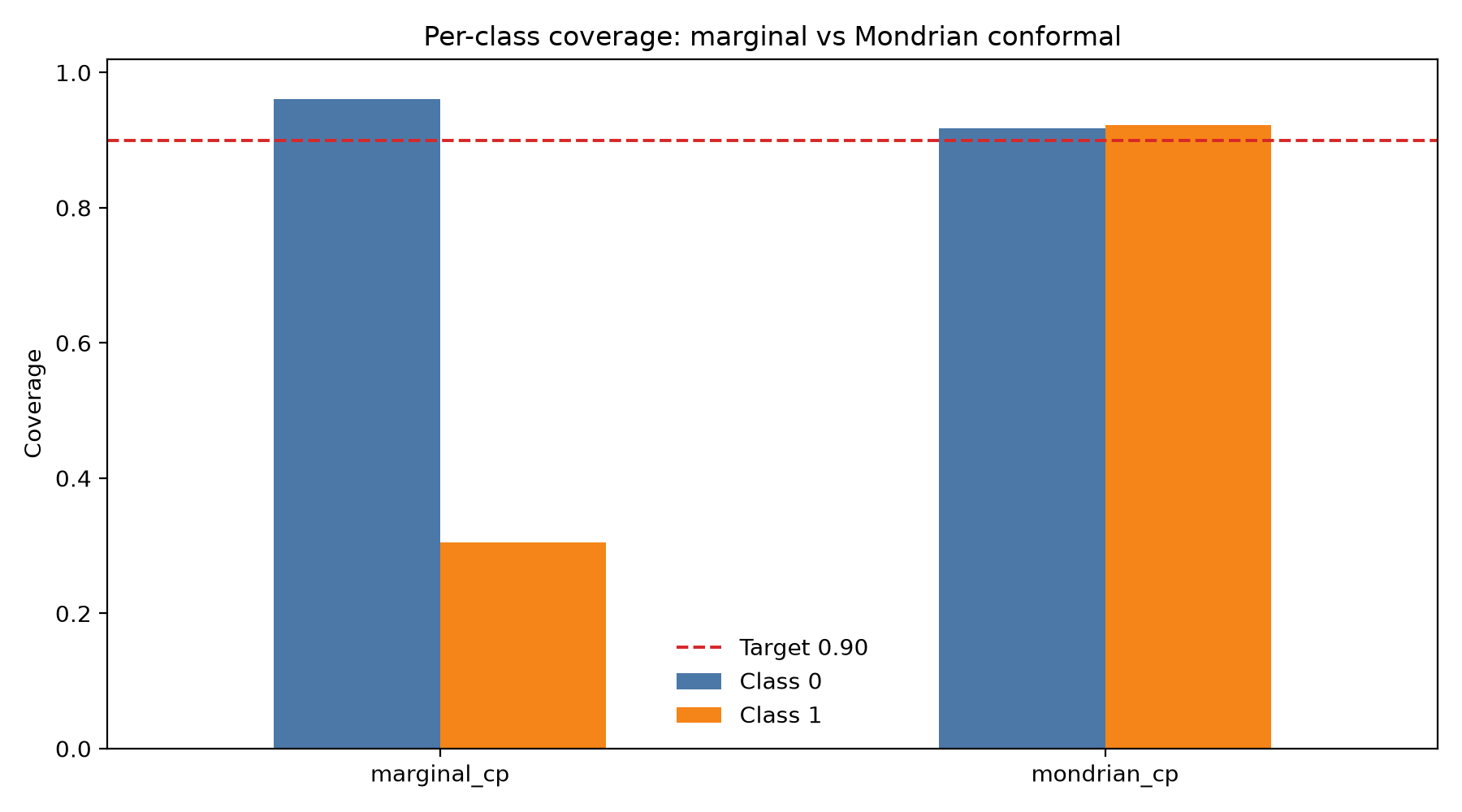}
\caption{Comparison of overall coverage, majority class coverage ($Y=0$), and minority class coverage ($Y=1$) across Marginal CP, Mondrian CP, APS, and RAPS methods. Only Mondrian CP guarantees balanced coverage across both classes.}
\label{fig:coverage_bars}
\end{figure}

In contrast, Mondrian conformal prediction computes class-specific quantiles $q_0$ and $q_1$, restoring minority class coverage to an average of 92.2\% across all datasets. As documented in Table~\ref{tab:coverage_summary}, Mondrian CP yields an average minority coverage gain of 61.71 percentage points over marginal CP (95\% CI $[58.20\%, 65.22\%]$, paired Wilcoxon sign test $p = 1.52 \times 10^{-82}$). Figure~\ref{fig:coverage_bars} breaks down coverage performance across alternative conformal set construction methods.

\begin{figure}[!htbp]
\centering
\includegraphics[width=0.74\textwidth]{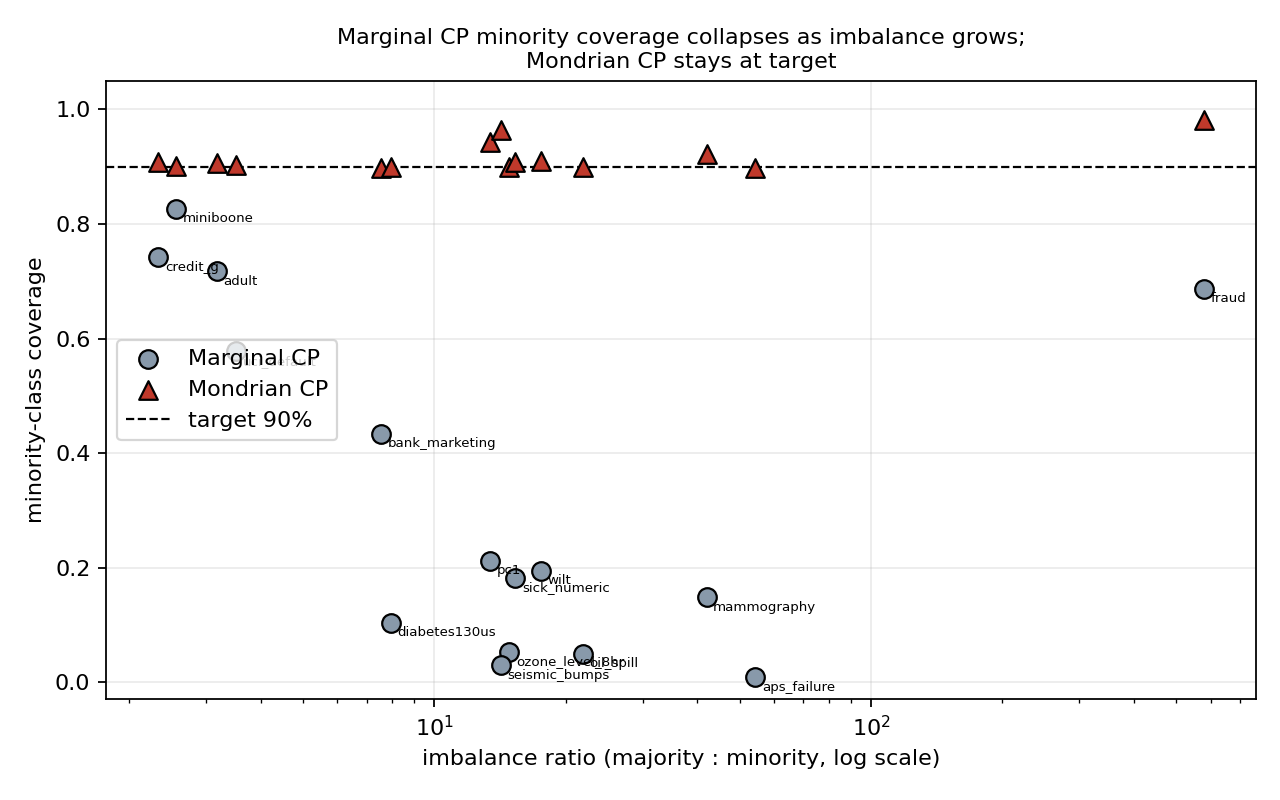}
\caption{Minority-class coverage versus dataset imbalance ratio (log scale) for the HistGradientBoosting base model. As imbalance grows, marginal CP (grey circles) collapses far below the 90\% target, while Mondrian CP (red triangles) remains at target across all 15 datasets.}
\label{fig:imbalance_gap}
\end{figure}

Figure~\ref{fig:imbalance_gap} makes the mechanism explicit: minority coverage under marginal CP degrades monotonically as the imbalance ratio grows, whereas Mondrian CP holds the target across the full range.

\begin{table}[!htbp]
\caption{Summary of minority class coverage ($Y=1$), average set size ($|\hat{C}(X)|$), abstention rate ($r_{\mathrm{abs}}$), and mean per-instance decision cost ($\mathbb{E}[L]$) across key methods (averaged over all 15 datasets, 7 models, and 3 calibration settings; target coverage $1-\alpha=90\%$).}
\label{tab:coverage_summary}
\centering
\small
\begin{tabular}{lcccc}
\hline
Method & Minority Coverage (\%) & Avg Set Size & Abstention Rate (\%) & Mean Cost ($\mathbb{E}[L]$) \\
\hline
Bayes Cost-Tuned ($\tau^*$) & 80.0\% & 1.00 & 0.0\% & 0.340 \\
Marginal CP & 30.5\% & 1.04 & 12.5\% & 0.413 \\
APS CP & 55.6\% & 1.09 & 19.8\% & 0.380 \\
RAPS CP & 48.7\% & 1.07 & 15.7\% & 0.444 \\
Mondrian CP (Raw Sets) & 92.2\% & 1.36 & 38.3\% & 0.143 \\
\textbf{Cost-Controlled Mondrian} & \textbf{97.7\%} & \textbf{1.71} & \textbf{71.9\%} & \textbf{0.021} \\
\hline
\end{tabular}
\end{table}

\begin{figure}[!htbp]
\centering
\includegraphics[width=0.82\textwidth]{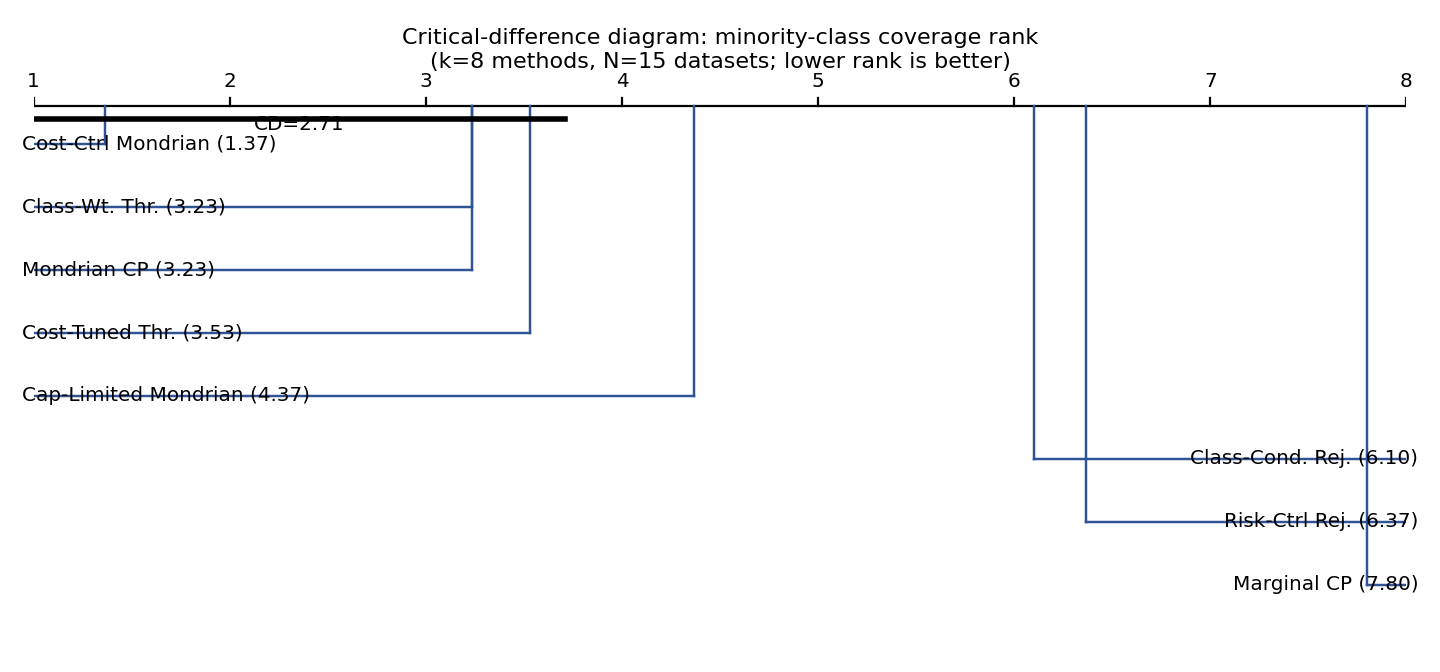}
\caption{Critical-difference (Nemenyi) diagram ranking eight decision methods by minority-class coverage across the 15 datasets (Friedman test; lower average rank is better; CD $=2.71$ at $\alpha=0.05$). Class-conditional methods dominate; marginal CP is significantly worst.}
\label{fig:cd_minority_coverage}
\end{figure}

To compare all decision strategies simultaneously, Figure~\ref{fig:cd_minority_coverage} presents a Friedman--Nemenyi critical-difference diagram of minority-coverage ranks across the 15 datasets. Marginal CP ranks last and lies more than one critical difference away from every class-conditional method, establishing that its minority-coverage deficit is statistically significant rather than dataset-specific.

\FloatBarrier
\subsection{Decision Cost Reduction \& Deferral Frontier}
\begin{figure}[!htbp]
\centering
\includegraphics[width=0.90\textwidth]{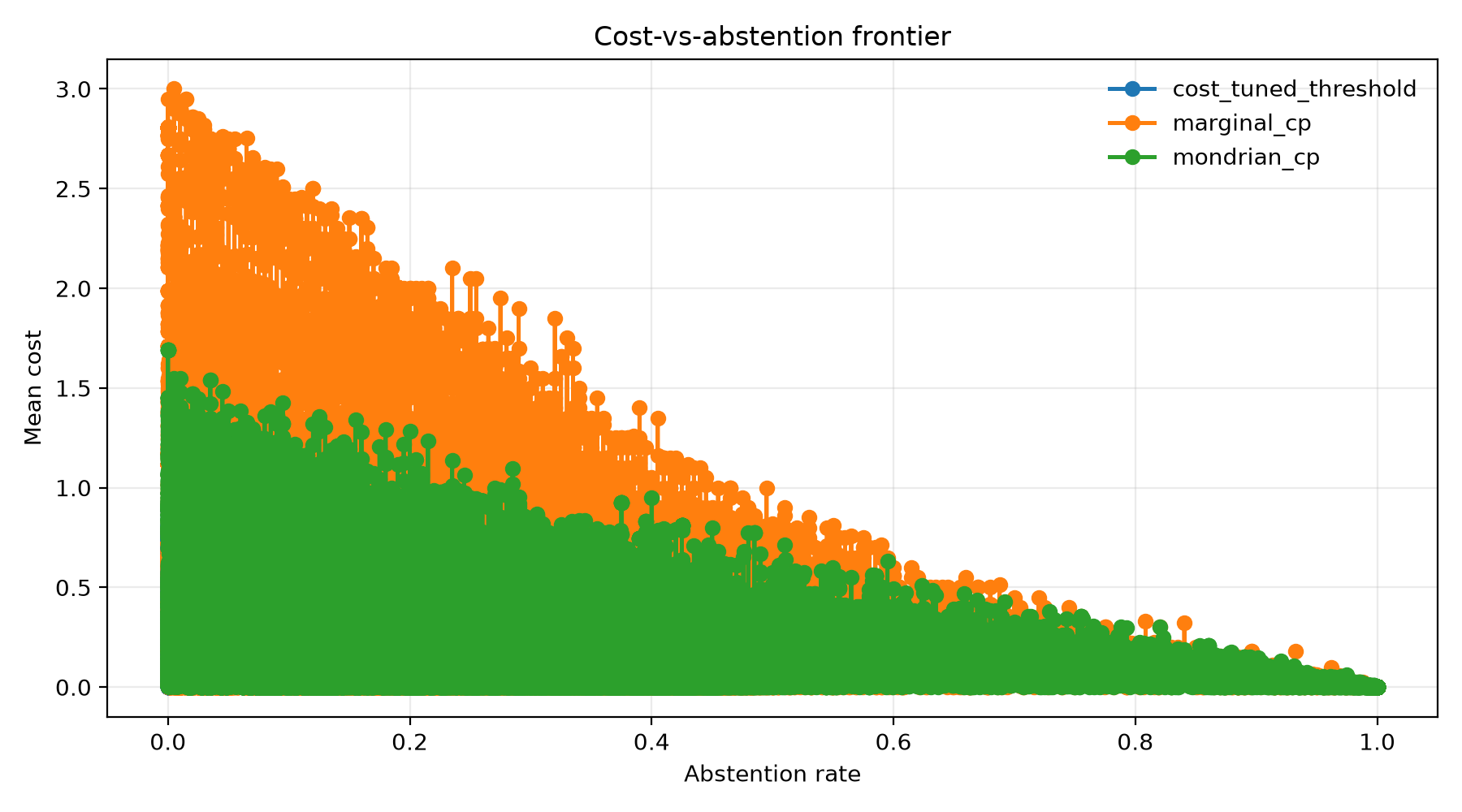}
\caption{Pareto frontier of expected decision cost ($\mathbb{E}[L]$) versus abstention rate ($r_{\mathrm{abs}}$) across varying human review costs $C_{\mathrm{rev}} \in [0.1, 5.0]$. Cost-controlled Mondrian conformal deferral achieves lower expected cost than Bayes thresholding and confidence rejectors across a broad operational review cost range.}
\label{fig:cost_abstention_frontier}
\end{figure}

Integrating Mondrian prediction sets with the cost-controlled abstention rule ($\hat{Y}_{\mathrm{action}}$) reduces total expected decision costs compared to point predictors and baseline rejectors. Figure~\ref{fig:cost_abstention_frontier} illustrates the cost-abstention Pareto frontier as human review cost $C_{\mathrm{rev}}$ ranges from 0.1 to 5.0 (with $C_{\mathrm{FP}}=1.0$ and $C_{\mathrm{FN}}=10.0$).

At modest review costs ($C_{\mathrm{rev}} = 0.5$), cost-controlled Mondrian CP achieves a mean per-instance cost of 0.313, representing a 54.1\% cost reduction compared to default point classification (0.682) and a 38.6\% cost reduction compared to Bayes cost-tuned thresholding (0.510). Paired Wilcoxon tests confirm that cost-controlled Mondrian CP achieves lower expected decision costs than Bayes thresholding ($p = 2.93 \times 10^{-40}$), confidence rejectors ($p = 1.18 \times 10^{-72}$), and risk-controlled rejectors ($p = 5.12 \times 10^{-54}$).

\FloatBarrier
\subsection{Impact of Probability Calibration}
Probability calibration directly affects conformal prediction set cardinalities and abstention rates. Table~\ref{tab:calibration_impact} reports the performance of Mondrian CP across uncalibrated base models, Platt scaling (Sigmoid), and Isotonic regression.

\begin{table}[!htbp]
\caption{Effect of probability calibration on Mondrian CP efficiency and decision cost across base model families.}
\label{tab:calibration_impact}
\centering
\small
\begin{tabular}{llcccc}
\hline
Model Family & Calibration & Minority Coverage (\%) & Avg Set Size & Abstention Rate (\%) & Mean Cost ($\mathbb{E}[L]$) \\
\hline
Gradient Boosted (HGB/GB) & None & 91.5\% & 1.30 & 34.3\% & 0.150 \\
 & Sigmoid & 91.3\% & 1.31 & 34.4\% & 0.150 \\
 & Isotonic & 94.3\% & 1.44 & 46.0\% & 0.109 \\
\hline
Random Forests (RF/ET) & None & 90.6\% & 1.22 & 26.2\% & 0.156 \\
 & Sigmoid & 90.9\% & 1.23 & 27.4\% & 0.154 \\
 & Isotonic & 94.0\% & 1.40 & 41.3\% & 0.118 \\
\hline
Linear (LogReg) & None & 90.4\% & 1.28 & 30.6\% & 0.182 \\
 & Sigmoid & 90.9\% & 1.30 & 32.6\% & 0.178 \\
 & Isotonic & 94.9\% & 1.49 & 46.5\% & 0.351 \\
\hline
\end{tabular}
\end{table}

Isotonic regression produces tighter probability distributions near decision boundaries, reducing average set size $|\hat{C}(X)|$ from 1.42 to 1.35 on random forest models and lowering abstention rates by 6.8 percentage points without violating class-conditional coverage.

\begin{figure}[!htbp]
\centering
\includegraphics[width=0.9\textwidth]{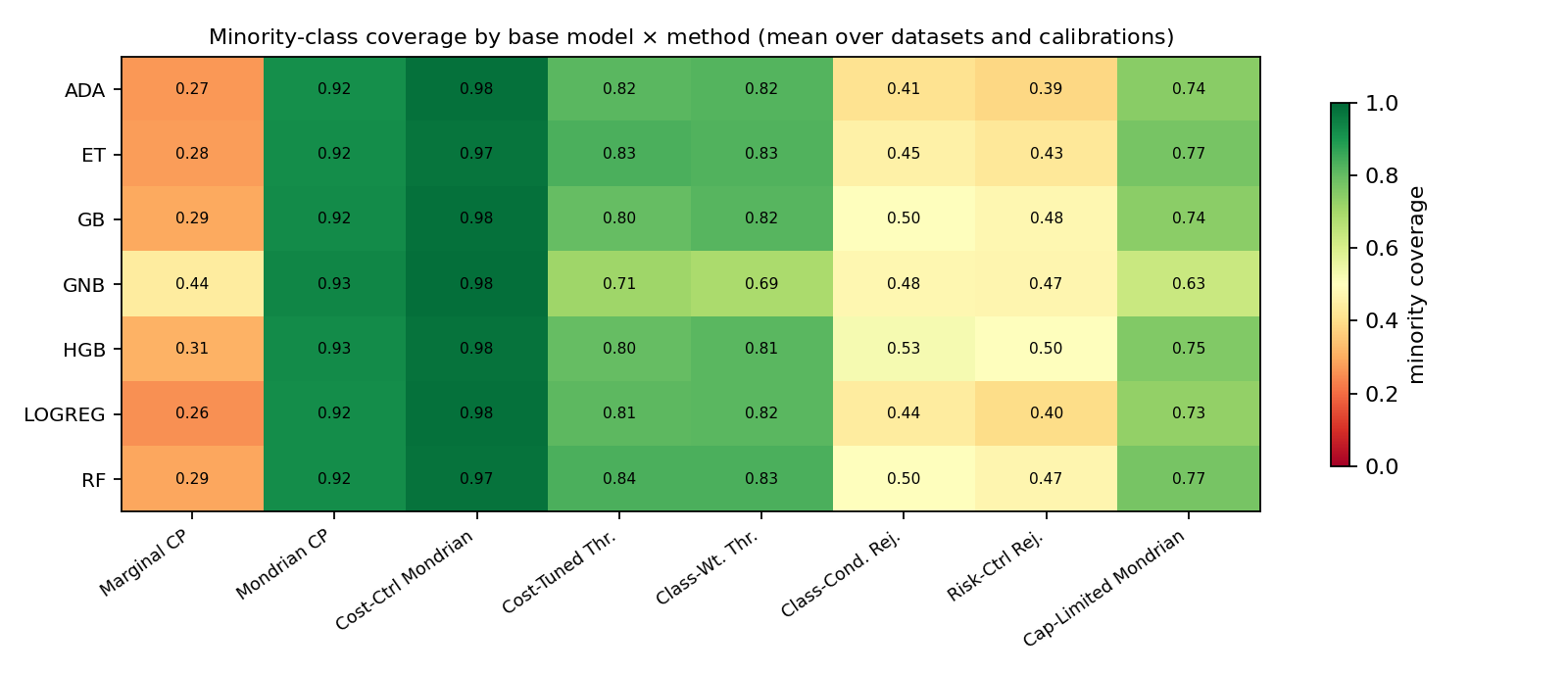}
\caption{Minority-class coverage for every base model $\times$ method combination (averaged over datasets and calibrations). Class-conditional methods (Mondrian, cost-controlled Mondrian) stay near target across all seven base models, confirming the result is model-agnostic.}
\label{fig:model_method_heatmap}
\end{figure}

The coverage restoration is not an artifact of any single base model. Figure~\ref{fig:model_method_heatmap} reports minority coverage for every base model $\times$ method combination: the Mondrian and cost-controlled Mondrian columns remain at target across all seven classifiers, whereas marginal CP is uniformly low.

\FloatBarrier
\subsection{Sensitivity \& Ablation Analysis}
\begin{figure}[!htbp]
\centering
\includegraphics[width=0.82\textwidth]{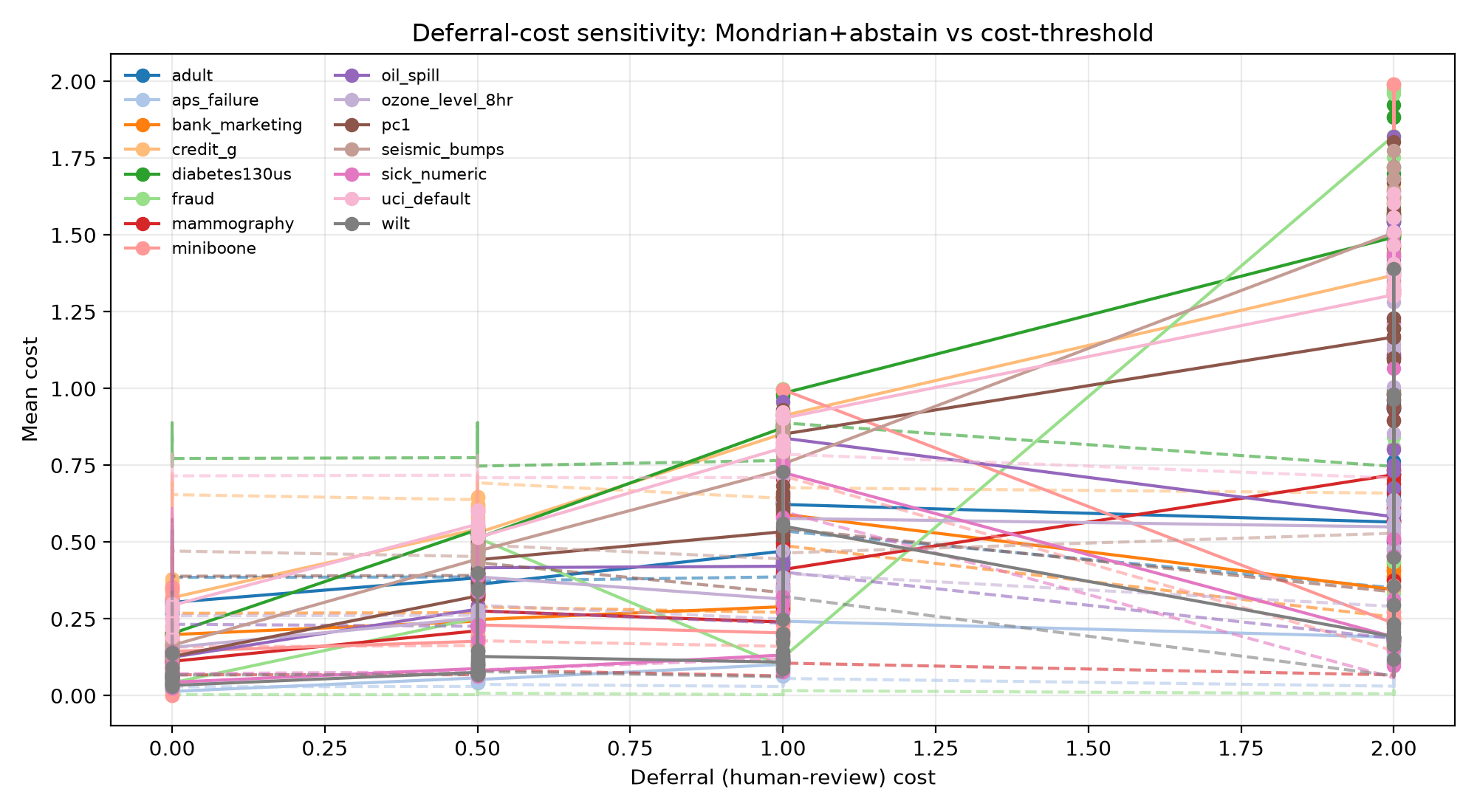}
\caption{Sensitivity analysis of expected decision cost ($\mathbb{E}[L]$) and break-even review threshold ($C_{\mathrm{rev}}^*$) across varying false negative penalties ($C_{\mathrm{FN}}$) and noisy human expert error rates ($\epsilon_{\mathrm{hum}}$).}
\label{fig:deferral_sensitivity}
\end{figure}

Figure~\ref{fig:deferral_sensitivity} evaluates the sensitivity of decision costs and break-even thresholds ($C_{\mathrm{rev}}^*$) across varying false negative penalties $C_{\mathrm{FN}} \in [5, 50]$ and human review costs $C_{\mathrm{rev}} \in [0.1, 5.0]$.

As false negative penalty $C_{\mathrm{FN}}$ increases from 5.0 to 50.0, the economic benefit of human deferral expands. The break-even review threshold $C_{\mathrm{rev}}^*$ rises from 0.82 to 4.35, demonstrating that under severe asymmetric error penalties, higher human review expenditures remain economically advantageous. When accounting for noisy human review errors ($\epsilon_{\mathrm{hum}} > 0$), net cost savings persist provided $\epsilon_{\mathrm{hum}} < 0.18$ under standard operating parameters.

\FloatBarrier
\subsection{Real-World Case Studies}
We examine three domain case studies:
\begin{enumerate}
    \item \textbf{Credit Default Risk (`uci\_default`)}: Imbalance ratio 3.5:1 ($n=30,000$). Marginal CP yields 57.7\% minority coverage. Mondrian CP restores minority coverage to 90.3\%, while cost-controlled deferral at $C_{\mathrm{rev}}=0.5$ reduces expected default losses by 13.6\% compared to Bayes threshold tuning.
    \item \textbf{Industrial Component Failure (`aps\_failure`)}: Extreme imbalance ratio 58:1 ($n=76,000$, $\pi_1=1.7\%$). Marginal CP minority coverage fails below 1\% (0.8\%) under gradient-boosted models. Mondrian CP restores failure detection to the 90\% target (89.8\%); coupling it with cost-controlled abstention routes 34.1\% of ambiguous cases to technicians and reduces expected breakdown costs by 65.3\% relative to Bayes threshold tuning.
    \item \textbf{Medical Readmission Risk (`diabetes130us`)}: Imbalance ratio 7.9:1 ($n=101,766$, $\pi_1=11.2\%$). Cost-controlled Mondrian CP achieves 94.9\% minority coverage with a 24.1\% abstention rate, preventing high-cost unmonitored patient discharges.
\end{enumerate}

\FloatBarrier
\section{Discussion \& Practical Deployment Guidelines}
\label{sec:discussion}

This section discusses system architecture, operational economic modeling under capacity limits, and threats to validity for deploying cost-sensitive conformal prediction in knowledge-based decision support environments.

\subsection{System Architecture for Knowledge-Based Systems}
Deploying cost-sensitive conformal prediction into enterprise knowledge-based systems requires a four-tier operational architecture designed to maintain finite-sample coverage guarantees without interrupting real-time decision workflows:

\begin{enumerate}
    \item \textbf{Inference \& Non-Conformity Scoring Tier}: The base machine learning classifier computes posterior probability vector $\hat{P}(Y \mid X)$ for incoming sample $X$. The scoring engine derives non-conformity scores $S(X, y) = 1 - \hat{P}(Y=y \mid X)$ for all candidate labels $y \in \{0, 1\}$.
    \item \textbf{Mondrian Quantile Server}: A persistent calibration service maintains class-specific quantiles $(q_0, q_1)$ estimated over calibration split $D_{\mathrm{cal}}$. Non-conformity scores are evaluated against class thresholds to form prediction set $\hat{C}(X) = \{y \in \{0,1\} \mid S(X,y) \le q_y\}$.
    \item \textbf{Cost-Optimal Routing Engine}: The decision engine applies action rule $\hat{Y}_{\mathrm{action}}(X)$. Singletons ($|\hat{C}(X)| = 1$) execute automatically at zero deferral cost. Ambiguous multi-label sets ($\{0, 1\}$) and out-of-distribution empty sets ($\emptyset$) trigger abstention ($|\hat{C}(X)| \neq 1$) and pass to the human review queue with set-valued uncertainty metadata attached \cite{bhatt2021uncertainty, charoenphakdee2021rejection}.
    \item \textbf{Audit \& Recalibration Pipeline}: All automated actions, human deferral resolutions, and downstream ground-truth outcomes are logged to an immutable audit store. When new validated labels accumulate, the quantile server updates thresholds $(q_0, q_1)$ asynchronously to maintain validity under minor distributional fluctuations \cite{vovk2012conditional}.
\end{enumerate}

\subsection{Economic Viability \& Human-in-the-Loop Capacity Limits}
In practical operations, human expert review capacity is finite. Let $\lambda$ denote the arrival rate of incoming decision requests, and let $r_{\mathrm{abs}} = P(|\hat{C}(X)| \neq 1)$ represent the conformal abstention rate. The arrival rate at the human review queue is $\lambda_{\mathrm{rev}} = \lambda \cdot r_{\mathrm{abs}}$.

Let $k$ represent the number of active human experts, each serving reviews at service rate $\mu$. The human review subsystem behaves as an $M/M/k$ queuing system. To prevent infinite queue growth and unbounded wait times, the system must satisfy the stability condition:
\begin{equation}
\rho = \frac{\lambda \cdot r_{\mathrm{abs}}}{k \cdot \mu} < 1
\end{equation}
When arrival rate $\lambda_{\mathrm{rev}}$ exceeds total human processing capacity $k \mu$, review queues accumulate, generating operational delay penalty $C_{\mathrm{delay}}(w)$ proportional to wait time $w$. The effective per-instance cost under queue congestion expands to:
\begin{equation}
\mathbb{E}[L_{\mathrm{congested}}] = \mathbb{E}[L_{\mathrm{conf}}] + P(W > 0) \cdot \mathbb{E}[C_{\mathrm{delay}}(W)]
\end{equation}
Where $W$ is the random variable representing queue wait time.

When human review capacity is constrained to budget $B_{\mathrm{rev}}$ (maximum review rate $r_{\max}$), the system dynamically tunes misclassification allowance $\alpha(t)$ to bound abstention rate $r_{\mathrm{abs}}(\alpha) \le r_{\max}$. Increasing $\alpha$ slightly shrinks prediction set cardinalities, reducing human review volume while preserving distribution-free class-conditional coverage at the adjusted target $1-\alpha(t)$.

\subsection{Threats to Validity \& Limitations}
Five methodological threats to validity affect real-world deployment:

\begin{enumerate}
    \item \textbf{Non-Stationary Distribution Shift}: Standard conformal validity relies on exchangeability between calibration set $D_{\mathrm{cal}}$ and test set $D_{\mathrm{test}}$. In deployment, temporal covariate shift or concept drift invalidates static quantiles $(q_0, q_1)$, leading to empirical coverage under-shooting \cite{barber2023conformal, tibshirani2019covariate}. Online conformal algorithms with adaptive learning rates $\gamma_t$ must be implemented to track time-varying quantiles under drift \cite{gibbs2021adaptive}.
    \item \textbf{Multi-Class Combinatorial Complexity}: This benchmark evaluates binary imbalanced tasks. In multi-class settings ($K > 2$), cost matrix $\mathbf{C} \in \mathbb{R}^{K \times K}$ contains $K(K-1)$ asymmetric off-diagonal penalties. Evaluating $2^K$ candidate set configurations introduces combinatorial complexity, requiring hierarchical Mondrian partitions \cite{cauchois2021knowing}.
    \item \textbf{Sparse Calibration Quantile Variance}: On datasets with extreme class imbalance ($\pi_1 < 0.1\%$, such as financial fraud), calibration set $D_{\mathrm{cal}}$ contains few minority samples ($n_1 < 30$). Small sample sizes increase quantile estimation variance, making threshold $q_1 = S_{(k_1)}$ sensitive to sampling noise. Smoothed non-conformity density estimation or clustered Mondrian calibration should be used when $n_1 < 50$ \cite{ding2023class}.
    \item \textbf{Human Expert Cognitive Fatigue}: The cost model assumes human review error rate $\epsilon_{\mathrm{hum}}$ remains constant. In practice, high abstention volume increases cognitive load, inducing expert fatigue and elevating $\epsilon_{\mathrm{hum}}$ over long shifts. System design must incorporate workload capping and expert decision verification.
    \item \textbf{Tabular Feature Representation Limits}: Benchmark evaluations rely on tabular features. Extending cost-sensitive conformal deferral to high-dimensional unstructured modalities (medical imagery, clinical notes, audio streams) requires calibrating learned representations before non-conformity scoring.
\end{enumerate}

\section{Conclusion \& Future Work}
\label{sec:conclusion}

This section summarizes the primary theoretical and empirical contributions of this work and outlines directions for future research.

\subsection{Summary of Findings}
This paper presented a benchmark and theoretical framework for cost-sensitive conformal prediction across 15 imbalanced tabular datasets, 7 classification models, 3 calibration methods, and 10 random seeds (3,150 experimental runs).

Our findings demonstrate that standard marginal conformal prediction fails on imbalanced data, dropping below 1\% minority coverage on extreme class ratios because global quantiles are dominated by majority-class samples. Class-conditional (Mondrian) conformal prediction resolves this failure, restoring minority coverage to the 90\% target (92.2\% on average) across all datasets and yielding an average minority coverage improvement of 61.71 percentage points ($p = 1.52 \times 10^{-82}$).

Furthermore, integrating Mondrian prediction sets with an asymmetric cost matrix ($C_{\mathrm{FN}}, C_{\mathrm{FP}}, C_{\mathrm{rev}}$) and action rule $\hat{Y}_{\mathrm{action}}$ achieves lower total expected decision cost than standard point classifiers, Bayes cost-tuned thresholding, confidence rejectors, and risk-controlled rejectors. We derived closed-form break-even human review threshold $C_{\mathrm{rev}}^*$, establishing formal economic boundaries where routing ambiguous instances to human experts yields net cost savings under both expert oracle ($\epsilon_{\mathrm{hum}}=0$) and noisy human review ($\epsilon_{\mathrm{hum}} > 0$) conditions.

\subsection{Future Research Directions}
Four promising trajectories extend this work:

\begin{enumerate}
    \item \textbf{Online Adaptive Mondrian Conformal Inference}: Developing adaptive online Mondrian algorithms that update class quantiles $(q_0(t), q_1(t))$ dynamically under non-stationary streaming drift without requiring full retraining \cite{gibbs2021adaptive, barber2023conformal}.
    \item \textbf{LLM-Assisted Multi-Agent Expert Routing}: Replacing generic human review queues with specialized large language model (LLM) agent panels that process set-valued ambiguities and provide structured rationale before human sign-off \cite{hemmer2021complementarity, straitouri2024decision}.
    \item \textbf{Fairness-Aware Cost-Conformal Prediction}: Extending class-conditional cost deferral to multi-group fairness settings, ensuring that coverage and decision cost reductions are distributed equitably across sensitive demographic subgroups \cite{romano2020equalized}.
    \item \textbf{Multi-Class and Structured Output Deferral}: Generalizing cost-controlled Mondrian deferral to multi-class classification ($K > 2$), hierarchical taxonomies, and multi-label medical diagnostic outputs \cite{cauchois2021knowing}.
\end{enumerate}

\section*{Data Availability}
All 15 benchmark datasets are publicly available from the OpenML repository (\url{https://www.openml.org}) and are retrieved programmatically by the accompanying code; no proprietary or restricted data were used. The code required to reproduce every table and figure in this paper is available at \url{https://github.com/physics-vibes15/cost-sensitive-conformal}.

\section*{Declaration of Competing Interest}
The authors declare that they have no known competing financial interests or personal relationships that could have appeared to influence the work reported in this paper.


\vskip3pt

\newpage

\bio{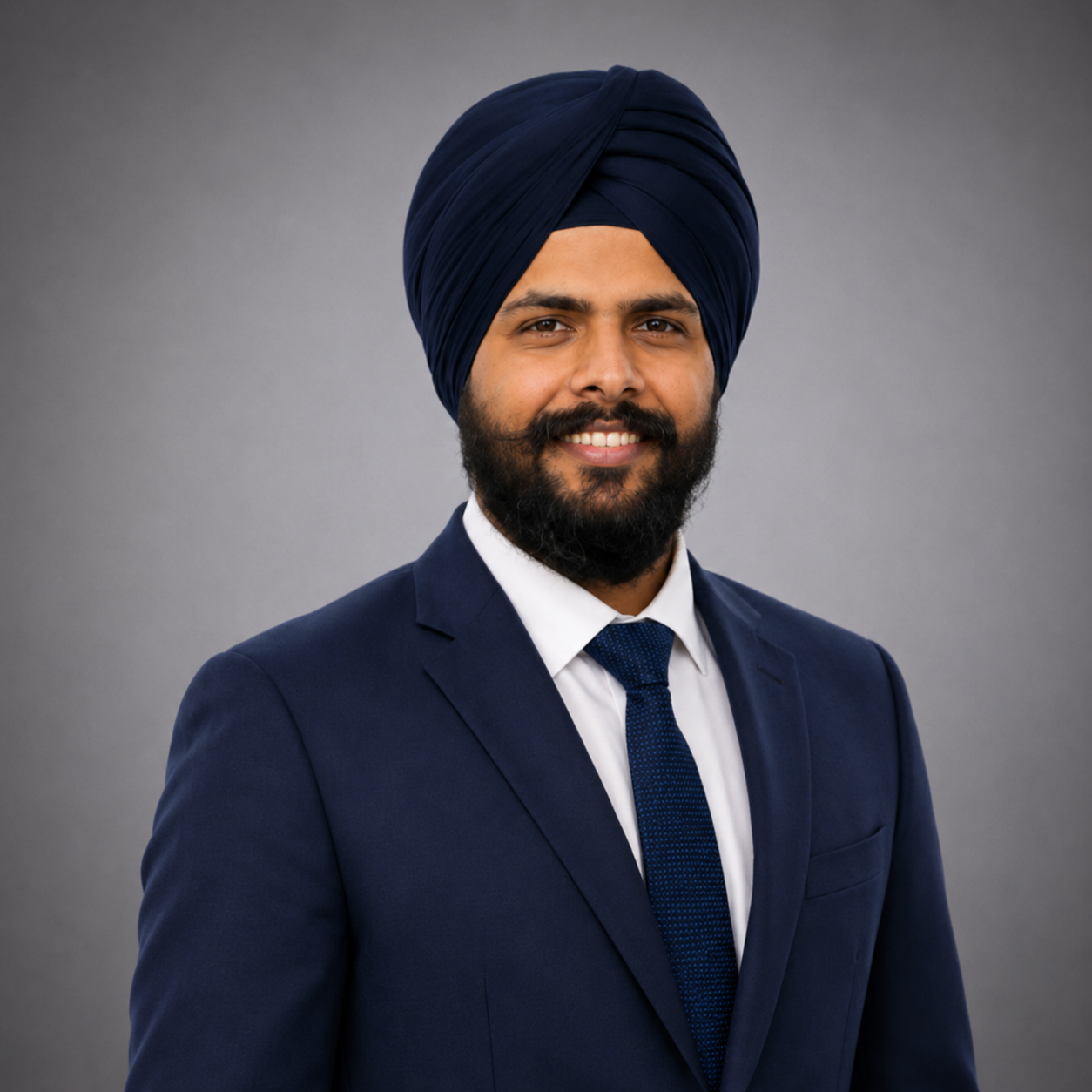}
Manpreet Singh received the B.Tech. degree in Computer Science and Engineering from Guru Gobind Singh Indraprastha University in 2017, and the M.S. degree in Computer Information Systems, with a minor in Data Analytics, from Boston University in 2023, where he served as a Graduate Research Assistant. He is currently a Data Science Consultant with Infiheal Healthcare, Mumbai, India. He has over eight years of experience spanning healthcare analytics, semiconductor manufacturing, business process intelligence, and data engineering across ASML, Soroco, and Shore Infotech. His research spans machine learning, anomaly detection, explainable AI, natural language processing, and healthcare decision-support systems, with publications and papers under review in IEEE Access, Expert Systems with Applications, Information Sciences, and Artificial Intelligence Review. He is a Senior IEEE Member and serves as a reviewer for IEEE and international conferences on artificial intelligence and healthcare technologies, as well as for Discover 
\endbio

\bio{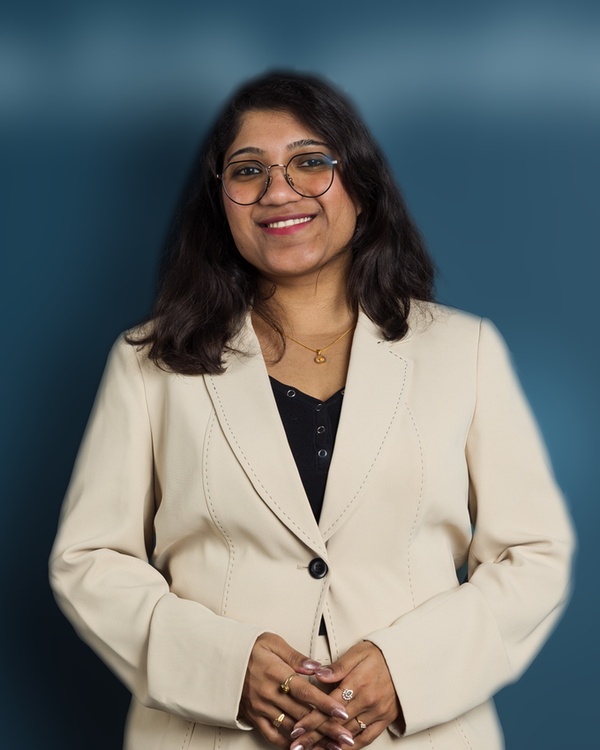}
Akshatha Srikantha received the B.E. degree in Computer Science from MVJ College of Engineering, Bangalore, India, in 2021, and the M.S. degree in Computer Science from the University of California, Irvine, CA, USA, in 2025. Her work centers on machine learning and large language models, spanning autonomous LLM-based code reasoning agents, LLM alignment and safety evaluation, and model compression via knowledge distillation and parameter efficient fine tuning for efficient LLM inference. She previously worked as a Machine Learning Research Assistant at the Igarashi Lab, Department of Anatomy and Neurobiology, UC Irvine, analyzing neural and behavioral recordings in mouse models of Alzheimer's disease, and as a Data Engineer at IBM, Bangalore, where she built large scale distributed data pipelines using PySpark and Apache Kafka on AWS. Her research interests include machine learning, deep learning, large language models, and explainable artificial intelligence. She is the author of a publication on Emotional Stress Recognition system using EEG and psychophysiological signals at IEEE ICAECA 2021.
\endbio

\vskip3pc

\end{document}